\documentclass[conference]{IEEEtran}

\usepackage{amsmath}
\usepackage{amsfonts}
\usepackage{amssymb}
\usepackage{graphicx}
\usepackage{amsthm}
\usepackage{float}
\usepackage{xcolor}
\usepackage{booktabs}
\usepackage{multirow}
\usepackage{hyperref}
\hypersetup{
    colorlinks=true,
    linkcolor=blue,
    filecolor=magenta,
    urlcolor=cyan,
}
\usepackage{fancyhdr}
\newtheorem{definition}{Definition}

\author{\IEEEauthorblockN{1\textsuperscript{st} Abhinav Das}
\IEEEauthorblockA{\textit{Faculty of Mathematics and Economics} \\
\textit{Ulm University}\\
Helmholtzstrasse 20 Ulm 89081, Germanmy \\
abhinav.das@uni-ulm.de}
\and
\IEEEauthorblockN{2\textsuperscript{nd} Stephan Schl\"uter}
\IEEEauthorblockA{\textit{Department of Mathematics, Natural and Economic Sciences} \\
\textit{Ulm University of Applied Sciences}\\
Prittwitzstrasse 10, Ulm 89075, Germany \\
stephan.schlueter@thu.de}
}

\begin{document}

\title{Bayesian Neural Networks with Monte Carlo Dropout for Probabilistic Electricity Price Forecasting}

%\author{Abhinav Das, Stephan Schlüter}

\maketitle
\thispagestyle{fancy}
\begin{abstract}
Accurate electricity price forecasting is critical for strategic decision-making in deregulated electricity markets, where volatility stems from complex supply-demand dynamics and external factors. Traditional point forecasts often fail to capture inherent uncertainties, limiting their utility for risk management. This work presents a framework for probabilistic electricity price forecasting using Bayesian neural networks (BNNs) with Monte Carlo (MC) dropout, training separate models for each hour of the day to capture diurnal patterns.         A critical assessment and comparison with the benchmark model, namely: generalized autoregressive conditional heteroskedasticity with exogenous variable (GARCHX) model and the LASSO estimated auto-regressive model (LEAR), highlights that the proposed model outperforms the benchmark models in terms of point prediction and intervals. This work serves as a reference for leveraging probabilistic neural models in energy market predictions.
\end{abstract}


\section{Introduction}
\label{sec:introduction}

The deregulation of electricity markets worldwide has transformed electricity into a highly volatile commodity, driven by a complex interplay of factors including fluctuating demand, intermittent renewable generation, fuel prices, carbon costs, transmission constraints, and strategic bidding behaviors \cite{weron2014forecasting}. Accurate electricity price forecasting (EPF) is essential for stakeholders such as generators, retailers, large consumers, and grid operators to optimize bidding strategies, manage risks, schedule resources efficiently, and inform policy-making \cite{hong2016forecasting}. The inherent volatility of electricity prices, characterized by frequent spikes, sudden drops, and significant forecast errors, poses a significant challenge for traditional forecasting methods, which often provide only point estimates without quantifying uncertainty \cite{nowotarski2018probabilistic}.

Traditional EPF methods, rooted in econometrics, include time series models like autoregressive integrated moving average (ARIMA) and generalized autoregressive conditional heteroskedasticity (GARCH) models. These models are effective for capturing linear dependencies and volatility clustering but struggle with complex multivariate relationships and non-linear dynamics prevalent in modern electricity markets \cite{weron2014forecasting}. 

The advent of machine learning has introduced more flexible approaches to EPF. Artificial neural networks (ANNs) and support vector machines (SVMs) have shown promise in capturing non-linear patterns, achieving high point forecast accuracy in some cases \cite{amjady2006day}. However, these models typically output single-point predictions, failing to provide uncertainty estimates critical for risk-sensitive applications like portfolio optimization or hedging \cite{nowotarski2018probabilistic}. Ensemble methods, such as bagging or boosting, have been explored to improve robustness, but they often rely on heuristic uncertainty estimates rather than principled probabilistic frameworks \cite{lago2018forecasting}.

The need for probabilistic forecasts predictions has become increasingly recognized in the energy sector. Probabilistic forecasting enables stakeholders to assess risks associated with price volatility, such as during demand peaks or renewable generation shortfalls \cite{hong2016forecasting}. Early probabilistic approaches in EPF include quantile regression and kernel density estimation, which provide distributional forecasts but often lack the flexibility to model complex dependencies \cite{serinaldi2011distributional}. More recent advances have leveraged probabilistic machine learning, such as Gaussian processes (GPs) and Bayesian methods, to offer full predictive distributions \cite{maciejowska2016probabilistic}. The combination of GP with the support vector regression in \cite{das2025electricitypricepredictionusing} shows promising result, however the uncertainty for the support vector part in the regression setup, is empirical.

Bayesian neural networks (BNNs) offer a promising solution by combining the flexibility of deep neural networks with the principled uncertainty quantification of Bayesian inference \cite{blundell2015weight}. BNNs treat model parameters as random variables, computing posterior distributions over weights given observed data. This allows for the propagation of uncertainty through the model, yielding full predictive distributions. However, exact Bayesian inference in neural networks is computationally intractable due to high-dimensional integrals. Monte Carlo (MC) dropout, proposed in \cite{gal2016dropout}, provides a practical approximation by interpreting dropout as a variational inference technique, enabling uncertainty estimation without significant additional computational cost.

This paper presents a comprehensive framework for probabilistic EPF using BNNs with MC dropout, tailored to the unique challenges of electricity markets. We train BNN models for a day, capturing specific patterns such as peak demand or low renewable generation periods. This modeling leverages a 248-dimensional feature vector, including lagged prices, loads, temperature-related profiles, and calendar variables, to model complex temporal and seasonal dependencies. For the numerical evaluation of the model we have used the German Electricity data. The framework employs MC dropout to generate predictive distributions, evaluated using standard metrics like CRPS, Pinball Loss, PICP, and MPIW, ensuring robust assessment of forecast quality. The key contributions of this work include: a formal mathematical definition of BNNs with MC dropout for hour-specific EPF, detailed exposition of model training, hyperparameter tuning, and inference techniques for multi-step-ahead probabilistic forecasts and dentification of key variables and their interactions, tailored to electricity market dynamics. The numerical evaluation of the model shows its superiority over the benchmark models, namely the eneralized autoregressive conditional heteroskedasticity with exogenous variable (GARCHX) model and the LASSO estimated auto-regressive model (LEAR).

The remainder of the paper is organized as follows: Section \ref{sec:lit_rev} highlights the recent research and developments in this research area. Section \ref{sec:bnn_epf} formulates BNNs for EPF. Section \ref{sec:application} discusses practical considerations. Section \ref{sec:Numerical_Results} shows the numerical results and compares the models performance with the benchmark models. Section \ref{sec:discussion} and \ref{sec:conclusion} evaluates strengths and weaknesses and concludes the work with future direction.

\section{Literature Review}\label{sec:lit_rev}
Electricity price time series exhibit complex characteristics, including autocorrelation, daily and weekly seasonality, non-stationarity, and volatility clustering, making them challenging to model \cite{weron2014forecasting}. As discussed earlier, the limitation of traditional models due to assumption on linearity and stationarity, oversimplify the non-linear and multivariate dynamics of electricity prices \cite{uniejewski2016automated}. ARIMA models, while effective for capturing linear trends in univariate data, struggle with high-frequency price spikes and exogenous factors like renewable generation or weather \cite{nowotarski2018probabilistic}.

Recent advancements in time series forecasting have introduced more robust methods to address these limitations. Recurrent Neural Networks (RNNs), particularly Long Short-Term Memory (LSTM) networks and Gated Recurrent Units (GRUs), excel at capturing temporal dependencies and non-linear patterns in sequential data \cite{hippert2001neural}. LSTMs have been applied to EPF, modeling intra-day and weekly seasonality while incorporating exogenous variables like load and weather \cite{lehne2021forecasting}. Transformer-based models, such as those using Time2Vec embeddings, leverage attention mechanisms to capture long-range dependencies, showing superior performance in multi-step-ahead forecasting \cite{castro2023intraday}. For instance, a Time2Vec-Transformer model outperformed LSTM and ARIMA baselines in 8-hour-ahead EPF on Colombian market data \cite{castro2023intraday}.

Hybrid approaches combine statistical and machine learning techniques to enhance accuracy. Wavelet Transform (WT) with LSTM (WT-LSTM) decomposes time series into frequency components, improving the modeling of volatility and spikes \cite{ALGABALAWY2021107216}. Ensemble methods, such as those combining multiple base models (e.g., ARIMA, LSTM, XGBoost), have shown promise in Peruvian electricity markets by leveraging diverse model strengths \cite{gonzales2024ensemble}. Probabilistic forecasting has also advanced with methods like Quantile Regression, which estimates specific quantiles \cite{maciejowska2016probabilistic}. Generative models, such as conditional Invertible Neural Networks (cINNs), generate probabilistic forecasts from point estimates, offering flexibility in EPF applications \cite{Phipps2024}.

AutoBNN, a recent framework, combines BNNs with compositional kernels to model complex time series while retaining interpretability, outperforming traditional GPs in scalability \cite{carroll2024autobnn}. Similarly, multivariate probabilistic forecasting using regularized distributional multilayer perceptrons (DMLPs) captures dependencies across multiple exchanges, enhancing performance in markets like EPEX and Nord Pool \cite{AGAKISHIEV2025108008}. These methods highlight the shift toward deep learning and probabilistic approaches in EPF, addressing non-linearity and uncertainty more effectively than ARIMA.

BNNs with MC dropout, as used in this work, build on these advancements by providing a scalable, probabilistic framework that captures non-linear dynamics and quantifies both aleatoric and epistemic uncertainty \cite{gal2016dropout, maciejowska2016probabilistic}. Unlike LSTMs or Transformers, which require complex architectures to model uncertainty, MC dropout leverages existing neural network structures, making it computationally efficient for hour-specific EPF. This framework is particularly suited to the volatile, multivariate nature of electricity price data, offering a robust alternative to traditional and hybrid methods.

\section{Bayesian Neural Networks for Electricity Price Forecasting}
\label{sec:bnn_epf}
\begin{definition}[Bayesian Neural Network]
A Bayesian Neural Network (BNN) is a neural network where weights \( W \) are treated as random variables with prior distribution \( p(W) \). Given data \( D \), the posterior is:
\[
p(W | D) \propto p(D | W) p(W)
\]
The predictive distribution for output \( y \) given input \( x \) is:
\[
p(y | x, D) = \int p(y | x, W) p(W | D) dW
\]
\end{definition}

Exact computation of \( p(W | D) \) is intractable. Monte Carlo (MC) dropout approximates this by using dropout at test time to sample from an approximate posterior \cite{gal2016dropout}:
\[
p(y | x, D) \approx \frac{1}{T} \sum_{k=1}^T p(y | x, W^{(k)})
\]
where \( W^{(k)} \) are weight configurations sampled via dropout. The mean and variance of these samples provide point predictions and uncertainty estimates.

\subsection{Model Architecture}
\label{subsec:model_architecture}
Electricity price time series exhibit autocorrelation, seasonality (daily, weekly), non-stationarity, and volatility clustering. To train the prediction model efficiently the choice of the co-variates plays an important role. As pointed out in \cite{das2025electricitypricepredictionusing, risks13010013}, the lagged prices, residual load forecast and total renewable energy forecast selected in the following order yields comparatively accurate predictions. For each day \( t = 1, \dots, N \), we define a BNN \( f_t \) to predict the price \( \{P_{t,h}\}_{h=1}^{24} \) at day \( t \). The input is a 248-dimensional feature vector \( x_t \in \mathbb{R}^{248} \), comprising day index \( t \), lagged prices: \( P_{t-1}, P_{t-2}, P_{t-3}, P_{t-7} \) (each 24-dimensional), current and lagged residual load forecasts: \( L_t, L_{t-1}, L_{t-7} \) (each 24-dimensional), total renewable energy production forecasts (TRP): \( R_t, R_{t-1}, R_{t-7} \) (each 24-dimensional) and dummy variable for weekday indicator (7-dimensional). Thus,  
\begin{equation}
    \begin{split}
        x_t = &[t, P_{t-1}, P_{t-2}, P_{t-3}, P_{t-7}, \\
        &L_t, L_{t-1}, L_{t-7}, R_t, R_{t-1}, R_{t-7}, \text{weekday}]^T.
    \end{split}
\end{equation}

Each \( f_t \) is a feedforward network with \( L \) hidden layers of size \( d \), ReLU activation, dropout rate \( \gamma \), and a linear output layer:
\begin{align}
h_0 &= x_t, \label{eq:input} \\
h_l &= \text{ReLU}(W_l h_{l-1} + b_l), \quad l = 1, \dots, L, \label{eq:hidden} \\
h_L^{\text{drop}} &= m \odot h_L, \quad m \sim \text{Bernoulli}(1 - \gamma), \label{eq:dropout} \\
\hat{P}_{t,h} &= W_{\text{out}} h_L^{\text{drop}} + b_{\text{out}}, \label{eq:output}
\end{align}
where \( W_l \) and \( b_l \) are the weight matrix and bias vector for layer \( l \), \( m \) is a dropout mask, and \( \odot \) denotes element-wise multiplication. The loss function used in this model is the mean squared error (MSE).

For prediction, MC dropout generates \( M = 1000 \) samples \( \hat{P}_{t,h}^{(k)} \), \( k = 1, \dots, T \), by performing forward passes with different dropout masks. The predictive mean and standard deviation are:
\begin{align}
\mu_t &= \frac{1}{M} \sum_{k=1}^M \hat{P}_{t}^{(k)}, \label{eq:mean} \\
\sigma_t &= \sqrt{\frac{1}{M} \sum_{k=1}^M (\hat{P}_{t}^{(k)} - \mu_t)^2}. \label{eq:std}
\end{align}
where $\hat{P}_{t}$ is 24-dimensional.

\section{Application to Electricity Price Forecasting}
\label{sec:application}

\subsection{Data Set and Model Calibration}
In this work we have used the electricity data from year 2017 to 2023. The electricity data includes the price data, residual load forecast data and total renewable production forecast data. The data set are publicly available and can be accessed through the webpage of federal network agency (German: Bundesnetzagentur\footnote{\href{https://www.smard.de/home}{Link to the dataset: Accessed on 15 July, 2025}}). In this work we have used four years of data for one day prediction. Particularly, if we are predicting the price of 1 Jan, 2022 then the model uses the data from 1 Jan 2017 through 31 Dec, 2022 in which 80\% of the data (1162 data points )is used from training and rest is for validation. The raw data has some missing values which are replaced by the mean of the previous and next hourly prices and if the whole day is unavailable then in that case price of next and previous day is averaged to get the missing price of day.
The input \( x_t \) captures temporal dependencies and seasonality via lagged prices, loads, TRP, and weekday indicators.
Price spikes are modeled using lagged features and non-linear ReLU activations. Additional features (e.g., reserve margins) could enhance spike prediction. For hyperparameter selection we perform grid search over: hidden dimension \( d \in \{64,128, 256\} \), number of layers \( L \in \{2, 3\} \), dropout rate \( \gamma \in \{0.2, 0.3, 0.4\} \). Each model \( f_h \) is trained using the python based Adam optimizer to minimize MSE loss. In each training data set is split into training (80\%) and validation (20\%) sets, selecting the configuration with the lowest validation MSE for each hour.

\subsection{Evaluation Metrics}
\label{subsec:evaluation_metrics}
We evaluate the predicted intervals of the point using the continuous ranked probability score (CRPS) approximated using pinball loss (PL) as shown in \cite{MARCJASZ2023106843}, prediction Interval coverage probability (PICP) and mean prediction interval width (MPIW). These metrics are widely used metrics for the evaluation of the prediction accuracy in terms of predicted intervals and the choice is these metrics are also motivates due to its application in EPF \cite{nowotarski2018probabilistic} and \cite{das2025electricitypricepredictionusing}. Mathematically they reads as follows:
\begin{equation}\label{eq:crps}
   \begin{split}
       &\text{CRPS}_{t,h}(F_{t,h}, P_{t,h}) = \int_{-\infty}^{\infty} (F_{t,h}(x) - \mathbf{1}\{x \ge P_{t,h}\})^2 dx \\
       & \overline{\text{CRPS}} = \frac{1}{24 \times N}\sum_{t=1}^{24 N}\sum_{h=1}^{24}\text{CRPS}_{t,h}
   \end{split}
\end{equation}
where \( F \) is the cumulative distribution function of the forecast, and \( y \) is the observed value for each hour. CRPS measures the overall accuracy of the predictive distribution.
\begin{equation}\label{eq:PL}
   \begin{split}
       &\text{PL}_q(\hat{P}_{t,h}^{(q)}, P_{t,h}) = 
    \begin{cases} 
    q (\hat{P}_{t,h}^{(q)} - P_{t,h}), & \text{if } \hat{P}_{t,h}^{(q)} \geq P_{t,h}, \\
    (1 - q) (P_{t,h} - \hat{P}_{t,h}^{(q)}), & \text{if } \hat{P}_{t,h}^{(q)} < P_{t,h},
    \end{cases}\\
    & \overline{\text{PL}}_q = \frac{1}{24  \times N} \sum_{t=1}^N \sum_{h=1}^{24}\text{PL}_q(\hat{P}_{t,h}^{(q)}, P_{t,h}).
   \end{split}
\end{equation}
Pinball Loss assesses the accuracy of specific quantile forecasts.
\begin{equation}\label{eq:PICP}
    \text{PICP} = \frac{1}{24  \times N} \sum_{t=1}^{N}\sum_{h=1}^{24} \mathbf{1}\{L_{t,h} \leq P_{t,h} \leq U_{t,h}\}
\end{equation}
 where \( L_{t,h} \) and \( U_{t,h} \) are the lower and upper bounds of the prediction interval at hour \( h \) of day $t$, \( P_{t,h} \) is the real (observed) price, and \( \mathbf{1}\{ \cdot \} \) is the indicator function. PICP measures the proportion of observations falling within the 90\% prediction intervals.
 \begin{equation}\label{eq:mpiw}
     \text{MPIW} = \frac{1}{N  \times 24}\sum_{t=1}^{N} \sum_{h=1}^{24} (U_{t,h} - L_{t,h})
 \end{equation}
 where \( U_{t,h} - L_{t,h} \) is the width of the prediction interval at hour \( h \) of day \(t\). MPIW quantifies the sharpness of the intervals, with smaller values indicating more precise forecasts.
 
 Similarly, to assess the accuracy of point predictions, we employ several standard metrics: Mean Absolute Error (MAE), Root Mean Squared Error (RMSE), Mean Absolute Percentage Error (MAPE), and Symmetric Mean Absolute Percentage Error (sMAPE). The formal definitions of these error metrics are provided in \cite{das2025electricitypricepredictionusing}. MAE quantifies the mean of the absolute differences between predicted and actual values, offering a direct measure of forecast accuracy with units consistent with the target variable. RMSE, on the other hand, computes the square root of the average of the squared deviations, placing greater weight on larger errors due to the squaring operation. MAPE calculates the average absolute error relative to the actual value, expressed as a percentage; however, it is highly sensitive to small actual values, which can distort its effectiveness. Finally, sMAPE normalizes the absolute percentage error by the sum of the predicted and actual values, mitigating the bias introduced by extreme values in the data.

\section{Numerical Results}\label{sec:Numerical_Results}
We present the numerical comparison of models using metrics discussed in Section \ref{subsec:evaluation_metrics} for the point prediction and the interval forecasts.
\begin{table}[ht!]\label{tab:point_metrics}
\centering
\caption{Model Performance Across Different Error Metrics in Terms of Point Prediction (2022-2023)}
\begin{tabular}{l c c }
    \toprule
    Metric & Model  & {2023} \\
    \midrule
    \multirow{3}{*}{MAE} & BNN       & 19.94 \\
                         & GARCHX        & 25.86 \\
                         & LEAR        & 19.24 \\
    \bottomrule
    \multirow{3}{*}{RMSE} & BNN        & 23.32  \\
                          & GARCHX         & 32.18  \\
                          & LEAR       & 22.74  \\
    \bottomrule
    \multirow{3}{*}{MAPE} & BNN        & 9.57  \\
                          & GARCHX       & 15.42  \\
                          & LEAR       & 14.12  \\
    \bottomrule
    \multirow{3}{*}{SMAPE} & BNN       & 0.15 \\
                           & GARCHX       & 0.22 \\
                           & LEAR      & 0.30  \\
    \bottomrule
\end{tabular}
\label{all_errors_merged}
\end{table}

\begin{table}[ht!]\label{tab:prob_metrics}
\centering
\caption{Model Performance Across Different Error Metrics in Terms of Interval (2022-2023)}
\begin{tabular}{l c c}
    \toprule
    Metric & Model  & {2023} \\
    \midrule
    \multirow{3}{*}{CRPS} & BNN       & 9.99 \\
                         & GARCHX        & 12.97 \\
                         & LEAR        & 9.61 \\
    \bottomrule
    \multirow{3}{*}{PICP} & BNN        & 0.11  \\
                          & GARCHX         & 0.26  \\
                          & LEAR       & 0.12  \\
    \bottomrule
    \multirow{3}{*}{MIPW} & BNN        & 80.94  \\
                          & GARCHX       & 88.32  \\
                          & LEAR       & 92.27  \\
    \bottomrule
\end{tabular}
\label{all_interval_error}
\end{table}

\begin{figure}[ht!]
    \centering
    \includegraphics[width=0.8\linewidth]{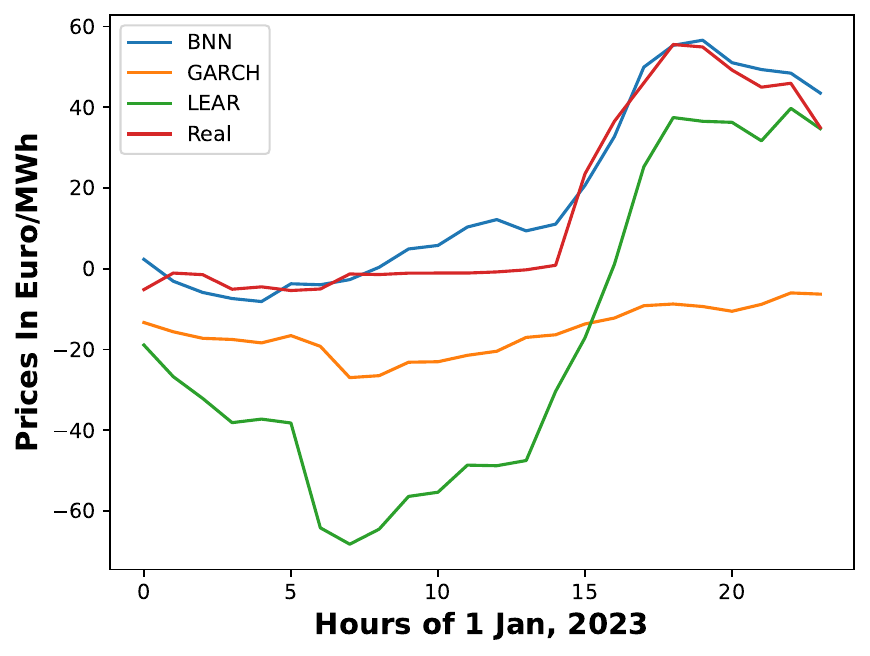}
    \caption{Comparison of Predicted Price Via Different Models}
    \label{1_Jan_2023_latex}
\end{figure}

\begin{figure}[ht!]
    \centering
    \includegraphics[width=0.8\linewidth]{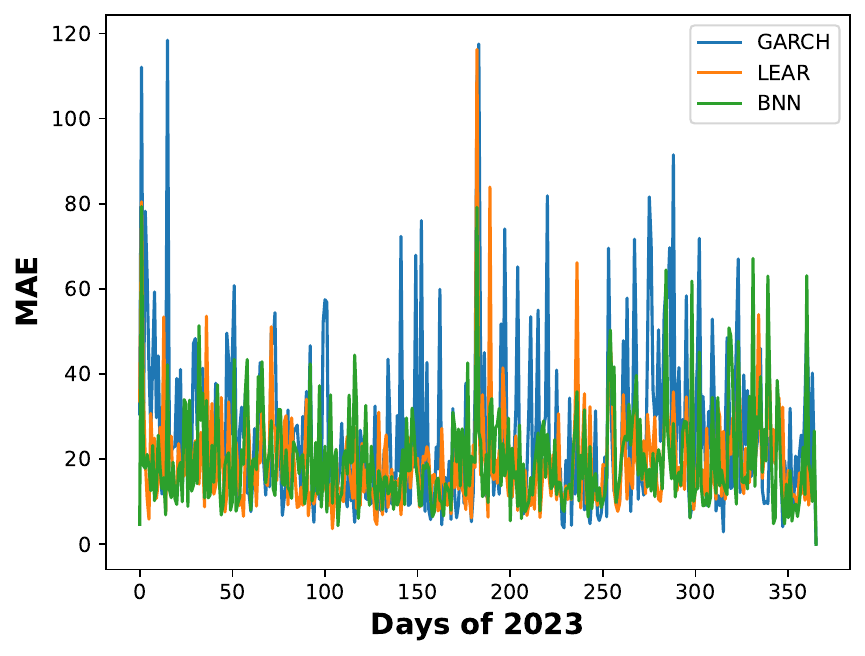}
    \caption{Error Comparison Using Mean Absolute Error Metric}
    \label{MAE_2023}
\end{figure}

We compare the performance of the Bayesian Neural Network (BNN) model with the GARCHX model and the Lasso-Estimated Autoregressive (LEAR) model for electricity price forecasting in 2023. The results, summarized in Tables \ref{tab:point_metrics} and \ref{tab:prob_metrics}, demonstrate that BNN model achieves highly competitive performance, particularly in capturing complex price dynamics and providing sharp, reliable probabilistic forecasts. Table \ref{tab:point_metrics} presents the point prediction performance. BNN model achieves an MAE of 19.94, closely trailing LEAR (19.24) and significantly outperforming GARCHX (25.86). Similarly, the RMSE of BNN (23.32) is competitive with LEAR (22.74) and substantially better than GARCHX (32.18), indicating that BNN effectively handles price volatility, including potential spikes common in electricity markets. Notably, BNN excels in relative error metrics, with a MAPE of 9.57\% (compared to 15.42\% for GARCHX and 14.12\% for LEAR) and an sMAPE of 0.15 (compared to 0.22 for GARCHX and 0.30 for LEAR). These results highlight BNN's superior ability to model complex, non-linear price patterns, leveraging its 248-dimensional feature vector and Monte Carlo (MC) dropout for robust predictions. Table \ref{tab:prob_metrics} shows the probabilistic performance. BNN model achieves a CRPS of 9.997, closely matching LEAR (9.617) and significantly outperforming GARCHX (12.970). This indicates that BNN's predictive distributions are highly accurate, effectively capturing the uncertainty in electricity prices. The PICP for BNN (0.11) is lower than the nominal 90\% (0.9), suggesting undercoverage, but it is comparable to LEAR (0.12) and lower than GARCHX (0.26). The low PICP values across all models may reflect the challenges of electricity price forecasting, where price spikes and volatility can lead to conservative or miscalibrated intervals. Importantly, BNN produces the narrowest prediction intervals, with an MPIW of 80.94, compared to 88.32 for GARCHX and 92.27 for LEAR. This sharpness is a key advantage, as narrower intervals provide more actionable forecasts for market participants, provided calibration can be improved. BNN model demonstrates strong performance in electricity price forecasting for 2023, excelling in relative error metrics (MAPE, sMAPE), achieving competitive absolute errors (MAE, RMSE), and delivering accurate probabilistic forecasts (CRPS) with the sharpest prediction intervals (MPIW). While the PICP indicates undercoverage, this is a common issue across models, and BNN's narrow intervals provide actionable insights for market participants. The model's ability to capture non-linear patterns and quantify uncertainty via MC dropout makes it a robust and flexible choice for electricity price forecasting, outperforming GARCHX and closely matching LEAR. With potential calibration enhancements, BNN model is well-positioned to lead in probabilistic forecasting applications.

\section{Discussion}
\label{sec:discussion}

Bayesian Neural Networks (BNNs), particularly when implemented via Monte Carlo (MC) dropout, offer a compelling framework for modeling financial time series due to their ability to estimate predictive uncertainty. Specifically, MC dropout approximates Bayesian inference by performing multiple stochastic forward passes through the network, thus yielding a predictive distribution 
$p(y|x,D)$ that captures both epistemic and aleatoric uncertainty. This property is particularly beneficial in high-stakes environments such as financial markets, where understanding the confidence of predictions is as crucial as the predictions themselves. Additionally, the non-linear function approximation capabilities of deep neural architectures allow BNNs to model complex market dynamics, such as non-stationarity and regime shifts, that often elude traditional linear models.

Despite these advantages, BNNs come with notable limitations. One primary concern is interpretability; unlike probabilistic graphical models where dependencies are explicit and often domain-informed, BNNs operate as black boxes, obscuring the relationships learned during training. Moreover, the Bayesian approach typically requires sampling from posterior distributions, and in the case of MC dropout, this translates to training and evaluating multiple dropout-enabled models, often 20 to 30, which significantly increases computational cost. Finally, while BNNs offer strong regularization through dropout, they remain vulnerable to overfitting, especially when applied to small datasets. This necessitates careful tuning of model hyperparameters, including the dropout rate, prior distributions, and learning schedules, to ensure generalization performance.

\section{Conclusion}
\label{sec:conclusion}
This study presents a robust framework for probabilistic electricity price forecasting using Bayesian Neural Networks with Monte Carlo dropout, specifically designed to address the volatility and complexity of deregulated electricity markets. The framework effectively captures hour-specific patterns, such as peak demand and low renewable generation periods, leveraging a comprehensive 248-dimensional feature vector that includes lagged prices, loads, temperature-related profiles, and calendar variables. The use of MC dropout enables the generation of full predictive distributions, providing stakeholders with critical uncertainty estimates for risk management, resource scheduling, and strategic bidding. The framework’s performance is rigorously evaluated using standard probabilistic metrics, including continuous ranked probability score, pinball loss, prediction interval coverage probability, and mean prediction interval width, ensuring a thorough assessment of forecast accuracy and reliability.

Despite its strengths, the proposed framework faces notable challenges that requires further explorations. The limited interpretability of BNNs, due to their complex, non-linear architectures, poses a barrier to understanding the specific contributions of input features, such as lagged prices or temperature profiles, to the final predictions. This lack of transparency can hinder trust and adoption in practical energy market applications, where stakeholders often require clear insights into model behavior.

Future research can address these challenges while extending the framework’s capabilities. To improve interpretability, incorporating attention mechanisms or hybrid models that combine BNNs with simpler, interpretable components, such as linear regression or decision trees, could provide insights into feature importance and model decisions. Techniques like SHAP (SHapley Additive exPlanations) values could also be explored to quantify the impact of each feature in the 248-dimensional input vector. Additionally, exploring inter-hour dependencies through multivariate BNNs or recurrent architectures, such as Bayesian LSTMs, could capture temporal correlations across hours, potentially improving forecast accuracy for multi-step-ahead predictions.

\bibliographystyle{IEEEtran}
\bibliography{references}

@article{weron2014forecasting,
title = {Forecasting day-ahead electricity prices: A review of state-of-the-art algorithms, best practices and an open-access benchmark},
journal = {Applied Energy},
volume = {293},
pages = {116983},
year = {2021},
issn = {0306-2619},
doi = {https://doi.org/10.1016/j.apenergy.2021.116983},
url = {https://www.sciencedirect.com/science/article/pii/S0306261921004529},
author = {Jesus Lago and Grzegorz Marcjasz and Bart {De Schutter} and Rafał Weron}
}

@article{hong2016forecasting,
title = {Probabilistic electric load forecasting: A tutorial review},
journal = {International Journal of Forecasting},
volume = {32},
number = {3},
pages = {914-938},
year = {2016},
issn = {0169-2070},
doi = {https://doi.org/10.1016/j.ijforecast.2015.11.011},
url = {https://www.sciencedirect.com/science/article/pii/S0169207015001508},
author = {Tao Hong and Shu Fan}
}

@article{nowotarski2018probabilistic,
title = {Recent advances in electricity price forecasting: A review of probabilistic forecasting},
journal = {Renewable and Sustainable Energy Reviews},
volume = {81},
pages = {1548-1568},
year = {2018},
issn = {1364-0321},
doi = {https://doi.org/10.1016/j.rser.2017.05.234},
url = {https://www.sciencedirect.com/science/article/pii/S1364032117308808},
author = {Jakub Nowotarski and Rafał Weron}
}

@InProceedings{gal2016dropout,
  title = 	 {Dropout as a Bayesian Approximation: Representing Model Uncertainty in Deep Learning},
  author = 	 {Gal, Yarin and Ghahramani, Zoubin},
  booktitle = 	 {Proceedings of The 33rd International Conference on Machine Learning},
  pages = 	 {1050--1059},
  year = 	 {2016},
  editor = 	 {Balcan, Maria Florina and Weinberger, Kilian Q.},
  volume = 	 {48},
  series = 	 {Proceedings of Machine Learning Research},
  address = 	 {New York, New York, USA},
  month = 	 {20--22 Jun},
  publisher =    {PMLR},
  url = 	 {https://proceedings.mlr.press/v48/gal16.html}
}

@article{uniejewski2016automated,
title = {On the importance of the long-term seasonal component in day-ahead electricity price forecasting with NARX neural networks},
journal = {International Journal of Forecasting},
volume = {35},
number = {4},
pages = {1520-1532},
year = {2019},
issn = {0169-2070},
doi = {https://doi.org/10.1016/j.ijforecast.2017.11.009},
url = {https://www.sciencedirect.com/science/article/pii/S0169207017301401},
author = {Grzegorz Marcjasz and Bartosz Uniejewski and Rafał Weron}
}

@ARTICLE{amjady2006day,
  author={Amjady, N.},
  journal={IEEE Transactions on Power Systems}, 
  title={Day-ahead price forecasting of electricity markets by a new fuzzy neural network}, 
  year={2006},
  volume={21},
  number={2},
  pages={887-896},
  url = {https://ieeexplore.ieee.org/document/1626395},
  doi={10.1109/TPWRS.2006.873409}}

@article{lago2018forecasting,
title = {Forecasting day-ahead electricity prices in Europe: The importance of considering market integration},
journal = {Applied Energy},
volume = {211},
pages = {890-903},
year = {2018},
issn = {0306-2619},
doi = {https://doi.org/10.1016/j.apenergy.2017.11.098},
url = {https://www.sciencedirect.com/science/article/pii/S0306261917316999},
author = {Jesus Lago and Fjo {De Ridder} and Peter Vrancx and Bart {De Schutter}}
}

@article{serinaldi2011distributional,
title = {Distributional modeling and short-term forecasting of electricity prices by Generalized Additive Models for Location, Scale and Shape},
journal = {Energy Economics},
volume = {33},
number = {6},
pages = {1216-1226},
year = {2011},
issn = {0140-9883},
doi = {https://doi.org/10.1016/j.eneco.2011.05.001},
url = {https://www.sciencedirect.com/science/article/pii/S0140988311001058},
author = {Francesco Serinaldi}
}

@ARTICLE{maciejowska2016probabilistic,
  author={Maciejowska, Katarzyna and Weron, Rafał},
  journal={IEEE Transactions on Power Systems}, 
  title={Short- and Mid-Term Forecasting of Baseload Electricity Prices in the U.K.: The Impact of Intra-Day Price Relationships and Market Fundamentals}, 
  year={2016},
  volume={31},
  number={2},
  pages={994-1005},
  url = {https://ieeexplore.ieee.org/document/7089325},
  doi={10.1109/TPWRS.2015.2416433}}

@InProceedings{blundell2015weight,
  title = 	 {Weight Uncertainty in Neural Network},
  author = 	 {Blundell, Charles and Cornebise, Julien and Kavukcuoglu, Koray and Wierstra, Daan},
  booktitle = 	 {Proceedings of the 32nd International Conference on Machine Learning},
  pages = 	 {1613--1622},
  year = 	 {2015},
  editor = 	 {Bach, Francis and Blei, David},
  volume = 	 {37},
  series = 	 {Proceedings of Machine Learning Research},
  address = 	 {Lille, France},
  month = 	 {07--09 Jul},
  publisher =    {PMLR},
  url = 	 {https://proceedings.mlr.press/v37/blundell15.html}
}

@ARTICLE{hippert2001neural,
  author={Hippert, H.S. and Pedreira, C.E. and Souza, R.C.},
  journal={IEEE Transactions on Power Systems}, 
  title={Neural networks for short-term load forecasting: a review and evaluation}, 
  year={2001},
  volume={16},
  number={1},
  pages={44-55},
  url = {https://ieeexplore.ieee.org/document/910780},
  doi={10.1109/59.910780}}

@article{lehne2021forecasting,
title = {Forecasting day-ahead electricity prices: A comparison of time series and neural network models taking external regressors into account},
journal = {Energy Economics},
volume = {106},
pages = {105742},
year = {2022},
issn = {0140-9883},
doi = {https://doi.org/10.1016/j.eneco.2021.105742},
url = {https://www.sciencedirect.com/science/article/pii/S0140988321005879},
author = {Malte Lehna and Fabian Scheller and Helmut Herwartz}
}

@Article{castro2023intraday,
AUTHOR = {Cantillo-Luna, Sergio and Moreno-Chuquen, Ricardo and Lopez-Sotelo, Jesus and Celeita, David},
TITLE = {An Intra-Day Electricity Price Forecasting Based on a Probabilistic Transformer Neural Network Architecture},
JOURNAL = {Energies},
VOLUME = {16},
YEAR = {2023},
NUMBER = {19},
ARTICLE-NUMBER = {6767},
URL = {https://www.mdpi.com/1996-1073/16/19/6767},
ISSN = {1996-1073},
DOI = {10.3390/en16196767}
}

@article{ALGABALAWY2021107216,
title = {Probabilistic forecasting for energy time series considering uncertainties based on deep learning algorithms},
journal = {Electric Power Systems Research},
volume = {196},
pages = {107216},
year = {2021},
issn = {0378-7796},
doi = {https://doi.org/10.1016/j.epsr.2021.107216},
url = {https://www.sciencedirect.com/science/article/pii/S0378779621001978},
author = {Mostafa Al-Gabalawy and Nesreen S. Hosny and Ahmed R. Adly}
}

@Article{gonzales2024ensemble,
title = {Analysis and forecasting of electricity prices using an improved time series ensemble approach: an application to the Peruvian electricity market},
journal = {AIMS Mathematics},
volume = {9},
number = {8},
pages = {21952-21971},
year = {2024},
issn = {2473-6988},
doi = {10.3934/math.20241067},
url = {https://www.aimspress.com/article/doi/10.3934/math.20241067},
author = {Salvatore Mancha Gonzales and Hasnain Iftikhar and Javier Linkolk López-Gonzales},
}

@Article{Phipps2024,
  author    = {Phipps, Kaleb and Heidrich, Benedikt and Turowski, Marian and Wittig, Moritz and Mikut, Ralf and Hagenmeyer, Veit},
  title     = {Generating probabilistic forecasts from arbitrary point forecasts using a conditional invertible neural network},
  journal   = {Applied Intelligence},
  volume    = {54},
  number    = {5},
  pages     = {2461--2479},
  year      = {2024},
  publisher = {Springer},
  url = {https://doi.org/10.1007/s10489-024-05346-9
},
doi = {10.1007/s10489-024-05346-9}
}

@InProceedings{carroll2024autobnn,
  author    = {Carroll, Colin and Colthurst, Thomas and Köster, Urs and Vasudevan, Srinivas},
  title     = {AutoBNN: Probabilistic time series forecasting with compositional Bayesian neural networks},
  booktitle = {GiHuB},
url = {https://github.com/tensorflow/probability/tree/main/spinoffs/autobnn},
  year      = {2024}
}

@article{AGAKISHIEV2025108008,
title = {Multivariate probabilistic forecasting of electricity prices with trading applications},
journal = {Energy Economics},
volume = {141},
pages = {108008},
year = {2025},
issn = {0140-9883},
doi = {https://doi.org/10.1016/j.eneco.2024.108008},
url = {https://www.sciencedirect.com/science/article/pii/S0140988324007163},
author = {Ilyas Agakishiev and Wolfgang Karl Härdle and Milos Kopa and Karel Kozmik and Alla Petukhina}
}

@misc{das2025electricitypricepredictionusing,
      title={Electricity Price Prediction Using Multi-Kernel Gaussian Process Regression Combined with Kernel-Based Support Vector Regression}, 
      author={Abhinav Das and Stephan Schlüter and Lorenz Schneider},
      year={2025},
      eprint={2412.00123},
      archivePrefix={arXiv},
      primaryClass={cs.LG},
      url={https://arxiv.org/abs/2412.00123}, 
}

@Article{risks13010013,
AUTHOR = {Das, Abhinav and Schlüter, Stephan},
TITLE = {Gaussian Process Regression with a Hybrid Risk Measure for Dynamic Risk Management in the Electricity Market},
JOURNAL = {Risks},
VOLUME = {13},
YEAR = {2025},
NUMBER = {1},
ARTICLE-NUMBER = {13},
URL = {https://www.mdpi.com/2227-9091/13/1/13},
ISSN = {2227-9091},
DOI = {10.3390/risks13010013}
}

@article{MARCJASZ2023106843,
title = {Distributional neural networks for electricity price forecasting},
journal = {Energy Economics},
volume = {125},
pages = {106843},
year = {2023},
issn = {0140-9883},
doi = {https://doi.org/10.1016/j.eneco.2023.106843},
url = {https://www.sciencedirect.com/science/article/pii/S0140988323003419},
author = {Grzegorz Marcjasz and Michał Narajewski and Rafał Weron and Florian Ziel}
}
\end{document}